\documentclass{article}

\usepackage{arxiv}

\usepackage[utf8]{inputenc} 
\usepackage[T1]{fontenc}    
\usepackage{hyperref}       
\usepackage{url}            
\usepackage{booktabs}       
\usepackage{amsfonts}       
\usepackage{nicefrac}       
\usepackage{microtype}      
\usepackage{lipsum}
\usepackage{graphicx}
\usepackage{amsmath}
\usepackage[table]{xcolor}
\graphicspath{ {./images/} }

\title{Dynamic Manifold Hopfield Networks for Context-Dependent Associative Memory}

\author{
 Chong Li \\
  Fudan University\\
  \texttt{lichong23@m.fudan.edu.cn} \\
   \And
 Taiping Zeng\thanks{Corresponding author} \\
  Fudan University\\
  \texttt{zengtaiping@fudan.edu.cn} \\
  \And
 Xiangyang Xue\footnotemark[1]\\
  Fudan University\\
  \texttt{xyxue@fudan.edu.cn} \\
  \And
  Jianfeng Feng\footnotemark[1]\\
  Fudan University\\
  \texttt{jffeng@fudan.edu.cn} \\
}

\begin{document}
\maketitle

\begin{abstract}
Neural population activity in cortical and hippocampal circuits can be flexibly reorganized by context, suggesting that cognition relies on dynamic manifolds rather than static representations.
However, how such dynamic organization can be realized mechanistically within a unified dynamical system remains unclear.
Continuous Hopfield networks provide a classical attractor framework in which neural dynamics follow gradient descent on a fixed energy landscape, constraining retrieval within a static attractor manifold geometry.
Extending this approach, we introduce Dynamic Manifold Hopfield Networks (DMHN), continuous dynamical models in which contextual modulation dynamically reshapes attractor geometry, transforming a static attractor manifold into a context-dependent family of neural manifolds.
In DMHN, network interactions are learned in a data-driven manner, to intrinsically deform the geometry of its attractor manifold across cues without explicit context-specific parameterization.
As a result, in associative retrieval, DMHN achieve substantially higher capacity and robustness than classical and modern Hopfield networks: when storing $2N$ patterns in a network of $N$ neurons, DMHN attain reliable retrieval with an average accuracy of 64\%, compared with 1\% and 13\% for classical and modern variants, respectively.
Together, these results establish dynamic reorganization of attractor manifold geometry as a principled mechanism for context-dependent remapping in neural associative memory.
\end{abstract}

\keywords{Dynamic Manifold Hopfield Networks \and neural manifolds \and associative memory \and attractor dynamics}

\section*{Significance Statement}
The brain can reliably recall memories while flexibly adapting to changing contexts, yet how this flexibility and stability are jointly achieved remains unclear.
We show that this balance can arise from dynamic reorganization of neural memory manifolds governed by attractor dynamics on Hopfield energy landscapes.
By introducing DMHN, we demonstrate how a single dynamical system can transition from a single static attractor manifold to a family of context-dependent dynamic neural manifolds while preserving fundamental attractor dynamics.
This framework establishes dynamic manifold reorganization as a principled mechanism for associative memory, linking context-dependent neural population dynamics in computational neuroscience to adaptive memory architectures in artificial intelligence.

\section{Introduction}

Human cognition is increasingly understood to be organized around dynamic neural manifolds.
Large-scale neural recordings have revealed that population activity evolves on manifolds embedded within high-dimensional neural state space \cite{stringer2019highdimensional, gallego2017neural, cunningham2014dimensionality}.
Moreover, accumulating evidence suggests that the geometry of these manifolds can be dynamically reorganized by task demands or contextual signals, allowing the same neural population to support qualitatively different computations across contexts \cite{manifoldperception, remington2018flexible, langdon2023unifying, perich2025neural}.
These findings position nerual manifolds as a unifying conceptual framework across computational neuroscience and artificial intelligence \cite{dubreuil2016storing, mHC, achilli2025the, HiddenManifoldModel, podlaski2025high, Li2024Representations}.
However, a central mechanistic question remains unresolved: \emph{how can recurrent neural circuits dynamically reorganize manifold geometry across changing contexts while preserving reliable computation?}

\begin{figure*}[!t]
	\centering
	\includegraphics[width=\linewidth]{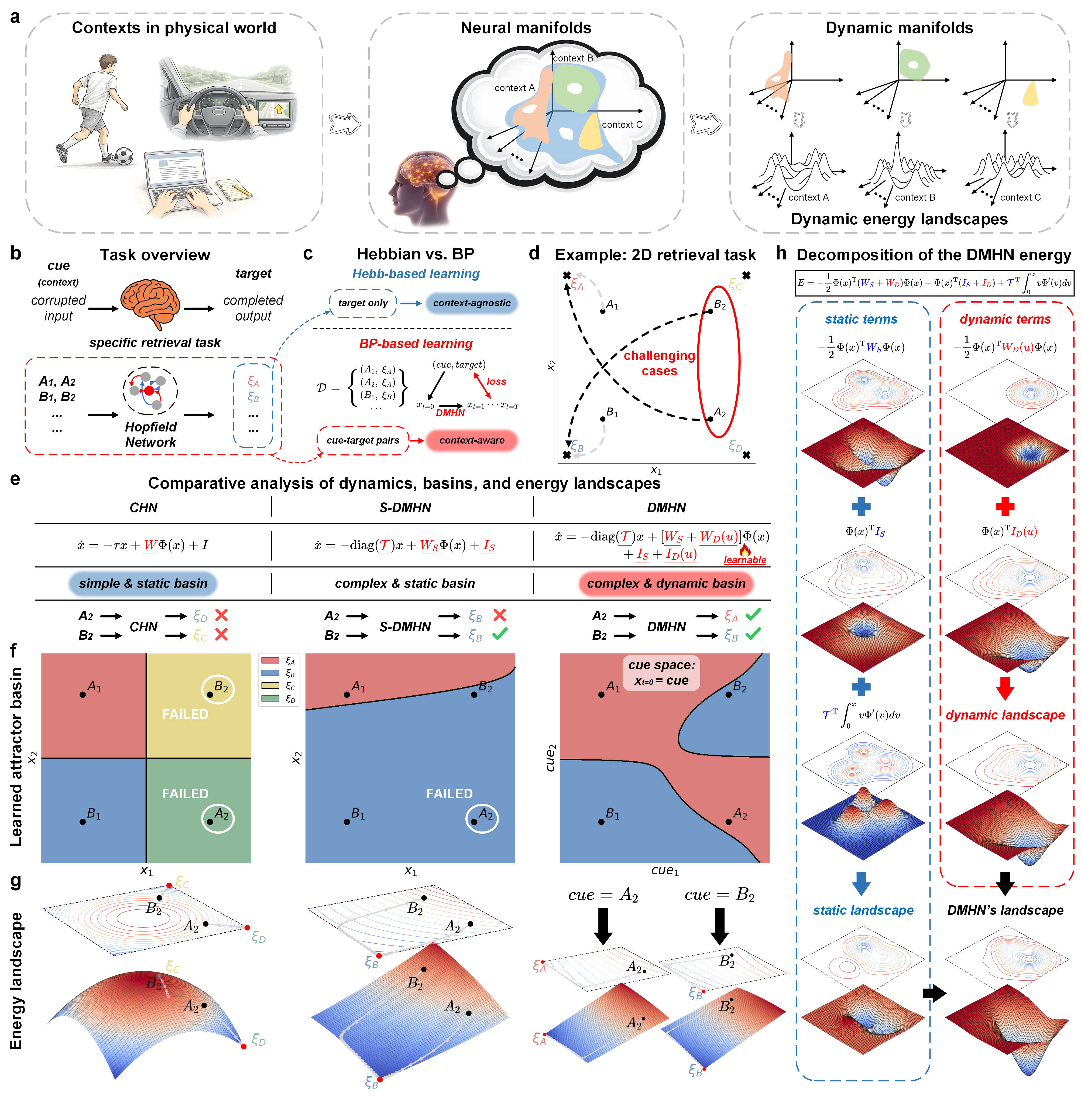}
	\caption{
		\textbf{Overview of Dynamic Manifold Hopfield Networks (DMHN).}
		(a) Experiences across contexts give rise to distinct neural manifolds \cite{perich2025neural}, suggesting that contextual changes reshape the underlying energy landscape and modulate attractor geometry.
		(b) Associative retrieval: a corrupted cue evolves through continuous attractor dynamics to produce a completed output, where the cue serves as the contextual signal.
		(c) Hebbian learning yields fixed attractor structures, whereas BP-based learning uses cue-target pairs to enable cue-conditioned retrieval.
		(d) A two-dimensional retrieval example illustrating challenging cue-attractor associations.
		(e,f) Comparison of CHN, S-DMHN, and DMHN in dynamics and learned basin geometry.
		(g) Corresponding energy landscapes: CHN and S-DMHN exhibit fixed landscapes, whereas DMHN induce cue-dependent effective landscapes.
		(h) Decomposition of DMHN energy into cue-independent (static) and cue-dependent (dynamic) terms, whose interaction generates dynamic manifold reorganization.
	}
	\label{fig:diagram}
\end{figure*}

Associative memory provides a canonical dynamical systems setting in which this question can be formulated.
It refers to the ability to retrieve complete representations from partial or noisy cues and constitutes a core function of both biological and artificial cognitive systems \cite{raaijmakers1981search, tulving1972, Kahana2008AssociativeRP}.
Importantly, memory retrieval is inherently dynamical at the level of neural population activity.
Rather than a direct mapping from input to output, neural activity evolves over time from a cue-dependent initial state toward a stable memory state, a process widely observed in cortical and hippocampal circuits \cite{rolls2018storage, rolls2024theory}.
These retrieval dynamics can be naturally interpreted as the convergence along manifolds embedded in neural state space, shaped by the structure of underlying memory attractors \cite{khona2022attractor}.

Attractor-based retrieval dynamics are classically captured by continuous Hopfield networks.
With symmetric recurrent synaptic interactions, Hopfield networks define an attractor system in which neural states evolve through a generalized gradient descent on an energy landscape, converging toward local minima that encode stored memories \cite{hopfield1982neural, hopfield1984neurons, halder2020hopfield}.
Such attraction strongly constrains neural dynamics, where activity rapidly collapses into low-energy basins, and state transitions preferentially follow directions associated with shallow energy barriers \cite{rolls2010attractor, betteti2023input}. 
As a result, neural population activity is locally confined to an implicit lower-dimensional manifold embedded in the state space, whose geometry is shaped by the underlying energy landscape \cite{khona2022attractor, MemoryManifolds}.
This formulation provides a principled account of pattern completion and has been closely linked to autoassociative memory circuits in hippocampal CA3, which are widely believed to support recall through attractor-like dynamics \cite{rolls2018storage, rolls2024theory}.
Consequently, Hopfield networks have long served as a foundational model for studying attractor-based associative memory in both neuroscience and artificial intelligence.

Despite these strengths, classical Hopfield networks have a well-known limitation in storage capacity.
Under standard assumptions of random pattern storage and Hebbian learning \cite{hebbian}, reliable retrieval is only possible for approximately $0.138N$ stored patterns in a network of $N$ neurons \cite{HopfieldCapacity}.
This limitation has motivated extensive efforts to increase the capacity of Hopfield-type memories while preserving their attractor dynamics.

One line of work addresses this limitation by modifying Hebbian learning rules.
Allowing the synaptic weight matrix to be an arbitrary symmetric matrix increases capacity to $N$ \cite{InformationCapacity}.
Within the Hebbian learning framework, further relaxing the zero-diagonal constraint yields a capacity of $CN\log N$ in the high-load regime \cite{HopfieldCapacityNew}.
Similar bounds have also been reported for continuous Hopfield networks \cite{ContinuousHopfieldCapacity}.
While these approaches improve storage limits, the resulting attractor geometry remains fixed, limiting the flexibility of retrieval dynamics across cues or contexts.

A complementary strategy increases capacity by altering the form of neuronal interactions.
Dense and modern Hopfield networks introduce higher-order interactions to dramatically expand storage capacity, in some cases achieving exponential scaling with the number of neurons \cite{krotov2016dense, demircigil2017model, ramsauer2020hopfield}.
These models also establish close connections between associative memory and modern deep learning architectures, such as attention mechanisms \cite{attention}.
However, the introduced many-body interaction structures depart from classical recurrent attractor dynamics and pose challenges for biological interpretability \cite{krotov2021largeassociativememoryproblem, kozachkov2025neuronastrocyte}.

\begin{figure*}[!t]
	\centering
	\includegraphics[width=0.99\linewidth]{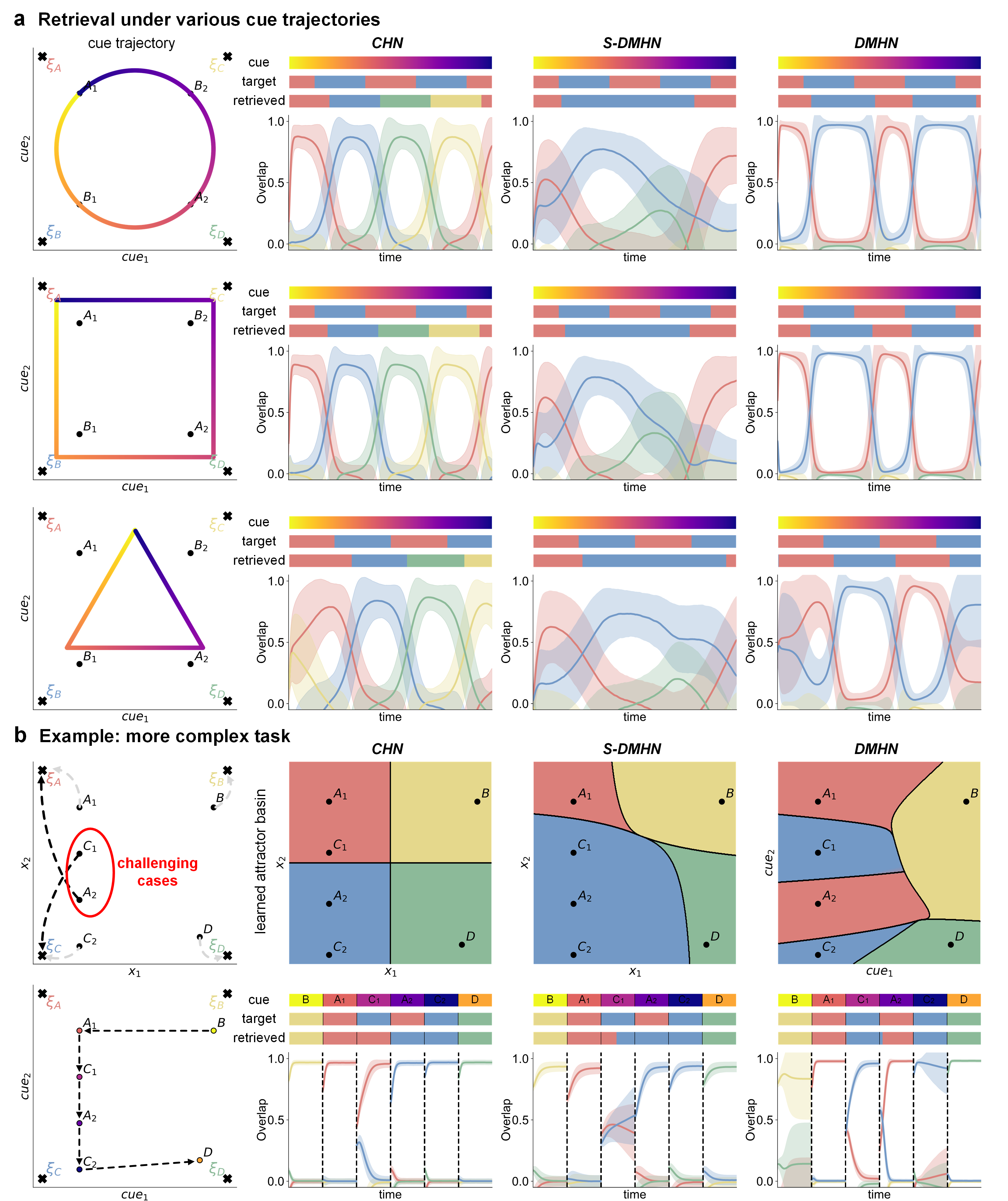}
	\caption{
		\textbf{Cue-dependent dynamic manifolds in DMHN.}
		(a) Network state trajectories under continuously varying cues. As the cue changes, population activity remains constrained to a low-dimensional manifold whose geometry is dynamically reshaped, demonstrating that retrieval dynamics are governed by cue-dependent manifold deformation rather than transitions among fixed attractors.
		(b) A more complex two-dimensional retrieval task, in which 6 distinct cues converge to 4 target attractors. Unlike (a), cues are corrupted and held fixed during retrieval, and network dynamics are initialized from the noisy cue at onset.
	}
	\label{fig:overlap}
\end{figure*}

Besides storage capacity, associative memory in biological neural circuits must also support dynamic reorganization.
This property is prominent in biological systems, where hippocampal place-field remapping shows that identical neuronal populations can encode distinct contexts, and affective or task-related signals further reshape population-level responses and decisions \cite{muller1987effects, wills2005attractor, tarcsay2025context, moita2003hippocampal, maren2013contextual, mante2013context}.
In parallel, manifold-based concepts have been introduced in artificial intelligence as representational or regularization principles \cite{manifoldmixup, Zhu_2018_CVPR, mHC, Li2024Representations}, and recent theoretical work has shown that manifold structure can emerge in recurrent networks through intrinsic dynamics \cite{sagodi2024back, pellegrino2025rnns, MemoryManifolds}.
Motivated by these developments, recent Hopfield-type models have begun to explore context-dependent modulation of associative memory.
For example, Betteti et al. \cite{betteti2023input} proposed an Input-Driven Plasticity (IDP) model that reshapes the energy landscape by dynamically reweighting stored patterns according to external input. However, this modulation depends on explicitly storing all memory patterns to compute saliency weights.
Similarly, Podlaski et al. \cite{podlaski2025high} introduced a context-modular memory network that selectively stabilize context-relevant memories via imposed neuronal and synaptic gating, at the cost of storing and applying additional context-specific gating matrices.
Taken together, existing approaches realize dynamic neural manifolds through explicit and enumerative parameterization for individual memories or contexts, leaving unresolved \emph{how an integrative dynamical system can intrinsically and continuously reshape its attractor manifold through its own dynamics across contexts}.

To address this challenge, we introduce \textbf{Dynamic Manifold Hopfield Networks (DMHN)}, which extend continuous Hopfield networks by allowing context to modulate network dynamics and induce context-dependent energy landscapes whose associated manifolds reorganize continuously (Fig.~\ref{fig:diagram}).
Unlike classical Hopfield models learned with Hebbian rules and operating on a fixed energy landscape \cite{hopfield1984neurons, HopfieldCapacity}, or modern Hopfield variants that rely on many-body interactions \cite{krotov2016dense, demircigil2017model, ramsauer2020hopfield}, DMHN learn interactions in a data-driven manner while preserving pairwise connectivity, and realize dynamic manifolds through intrinsic dynamics rather than enumerative parameterization \cite{betteti2023input, podlaski2025high}.
Mechanistically, DMHN reconcile flexibility and stability by dynamically reorganizing neural attractor manifolds under contextual modulation, while preserving continuous, attractor-based retrieval dynamics.

\begin{figure*}[!t]
	\centering
	\includegraphics[width=\linewidth]{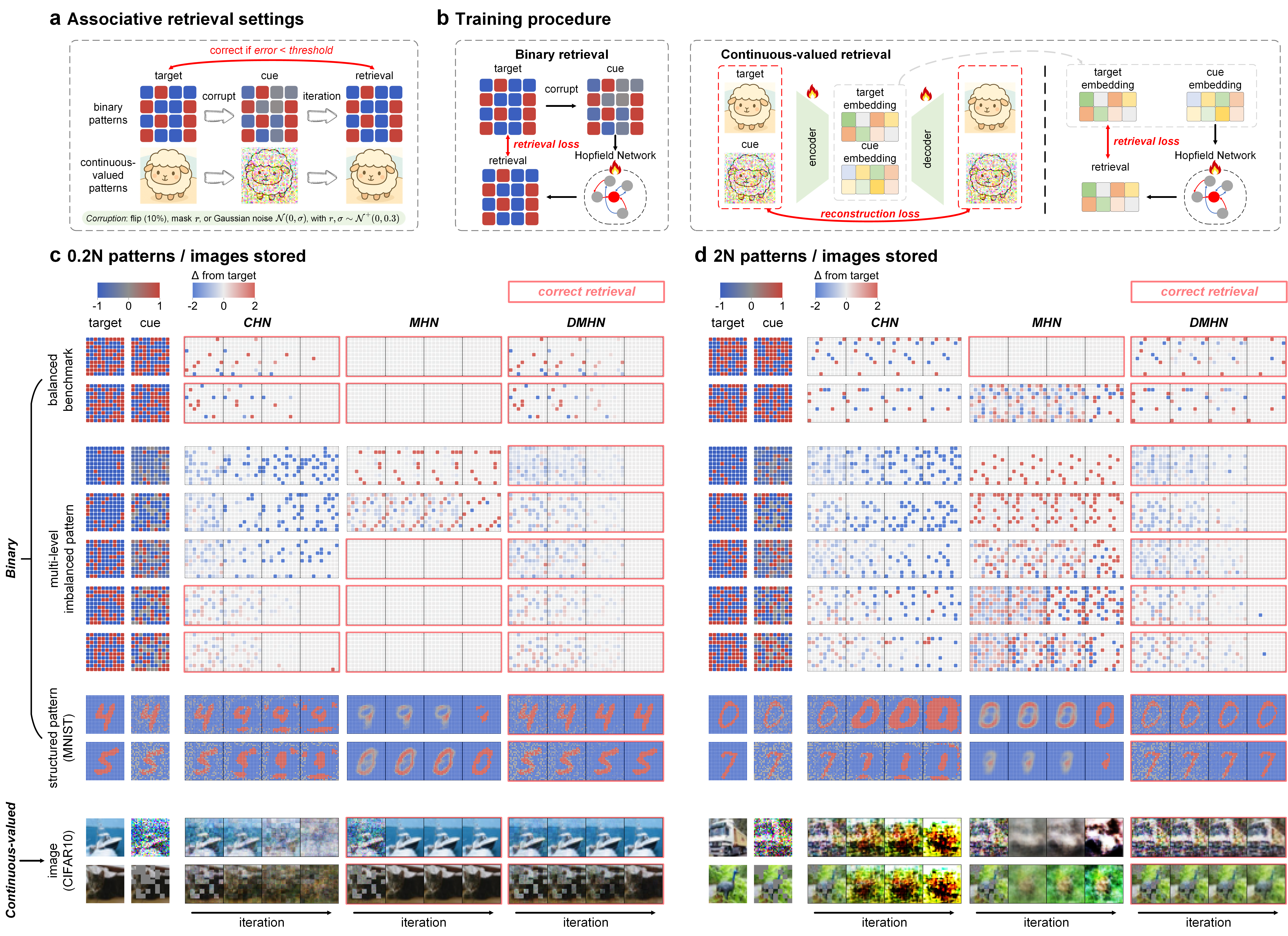}
	\caption{
		\textbf{Associative memory retrieval framework and dynamics.}
		(a) Associative retrieval settings.
		Binary and continuous-valued patterns are corrupted to generate cues, which are iteratively evolved toward target memories. 
		Corruption includes bit flips, masking, or additive Gaussian noise. 
		A retrieval is considered correct when the final error falls below a predefined threshold.
		(b) Training procedure.
		For binary retrieval, corrupted cues are iteratively evolved and a retrieval loss between the retrieved state and the target pattern is used to train the Hopfield network.
		For continuous-valued retrieval, target and cue embeddings are evolved in a shared latent space. Retrieval loss is applied between the retrieved embedding and the target embedding, while a reconstruction loss trains the encoder-decoder to reconstruct both targets and cues.
		(c,d) Retrieval dynamics under different memory loads.
		Representative retrieval trajectories for CHN, MHN, and DMHN when storing $0.2N$ patterns (c) and $2N$ patterns (d).
		For each dataset, a corrupted cue initializes the network, and successive states are shown across retrieval iterations.
		Under higher memory load, CHN and MHN exhibit increased distortion and convergence to mixed or spurious states, whereas DMHN trajectories remain more aligned with the target patterns.
		This trend is observed for both binary datasets (benchmark, imbalanced, and MNIST) and continuous-valued image representations (CIFAR10).
		Red boxes indicate retrievals considered correct according to the error threshold predefined in (a).
	}
	\label{fig:retrieval}
\end{figure*}

In numerical experiments, we evaluate DMHN across diverse associative memory settings and systematically compare DMHN with Classical Hopfield Networks (CHN) and Modern Hopfield Networks (MHN).
We first show that cue-conditioned modulation of the energy landscape in DMHN induces continuously deforming manifolds that shape retrieval trajectories, whereas retrieval in CHN and static DMHN unfold on a fixed landscape (Fig.~\ref{fig:diagram}, Fig.~\ref{fig:overlap}, SI Appendix, Movies S1--S4).
As a consequence, DMHN sustain reliable retrieval at substantially higher memory loads: when storing $2N$ patterns, DMHN achieve an average retrieval accuracy of 64\%, compared with 1\% for CHN and 13\% for MHN (Fig.~\ref{fig:retrieval}(c,d), Fig.~\ref{fig:pattern-results}(a--e), Tab.~\ref{tab:retrieval-results}).
Notably, this advantage persists in challenging regimes.
For highly imbalanced binary patterns, DMHN maintain 70\% accuracy for random patterns ($\frac{\#1}{\#-1}=\frac{1}{9}$) and 77\% for MNIST at $2N$ stored patterns, whereas both CHN and MHN fail to retrieve reliably (0\%).
This robustness further extends to continuous-valued memories derived from CIFAR10, where DMHN attain 84\% accuracy when storing $2N$ images, compared with 0\% for CHN and 18\% for MHN (Fig.~\ref{fig:pattern-results}(c,d), Tab.~\ref{tab:retrieval-results}).
Moreover, ablation analyses reveal that dynamic modulation of synaptic interactions ($W_D$) and biases ($I_D$) plays complementary roles in shaping retrieval dynamics under balanced and imbalanced conditions, respectively.
Overall, these results demonstrate that dynamic neural manifold reorganization provides a principled route to achieving high-capacity and robust associative memory while preserving continuous attractor-based retrieval dynamics.

\section{Results}

\subsection{Dynamic Manifold Hopfield Networks}

To interpret the following results, we first contrast the retrieval dynamics of Classical Hopfield Networks (CHN), Modern Hopfield Networks (MHN), and Dynamic Manifold Hopfield Networks (DMHN).
In CHN \cite{hopfield1984neurons}, the network state $\mathbf{x}\in\mathbb{R}^N$ evolves according to
\[
\dot{\mathbf{x}} = -\tau \mathbf{x} + \mathbf{W}\,\Phi(\mathbf{x}) + \mathbf{I},
\]
where retrieval dynamics unfold on a single fixed energy landscape.
Different retrieval cues influence the dynamics only through the initial condition $\mathbf{x}(0)$, leading trajectories to converge toward pre-existing attractors.

MHN \cite{ramsauer2020hopfield} achieve high storage capacity by replacing recurrent attractor dynamics with many-body, attention-like interactions.
Retrieval is typically expressed as an iterative update
\[
\mathbf{x}_{t+1}
=
\sum_{\mu=1}^{M}
\mathbf{y}^{\mu}\,
\frac{\exp\!\big(\beta \langle \mathbf{x}_t, \mathbf{k}^{\mu} \rangle\big)}
{\sum_{\nu=1}^{M}\exp\!\big(\beta \langle \mathbf{x}_t, \mathbf{k}^{\nu} \rangle\big)}.
\]
where learned key-value pairs $\{\mathbf{k}^{\mu}, \mathbf{y}^{\mu}\}$ determine the update.
While effective for high-capacity storage, this formulation is typically implemented as discrete iterative updates rather than continuous-time attractor flows.

DMHN preserve continuous attractor dynamics while allowing retrieval cues to modulate the effective retrieval geometry.
The network dynamics take the form
\[
\dot{\mathbf{x}}
=
-\mathrm{diag}(\boldsymbol{\mathcal T})\,\mathbf{x}
+
\big[\mathbf{W}_S + \mathbf{W}_D(\mathbf{u})\big]\Phi(\mathbf{x})
+
\mathbf{I}_S + \mathbf{I}_D(\mathbf{u}),
\]
where cue-independent and cue-dependent components jointly shape the dynamics.
For each fixed cue $\mathbf{u}$, DMHN define an autonomous attractor dynamics, while different cues induce systematically reorganized energy landscapes within a single dynamical system.

In the experiments reported below, the retrieval cue specifies both the initial condition and the corresponding retrieval manifold, while all network parameters remain fixed during evaluation.
This formulation enables direct examination of how cue-conditioned manifold reorganization differentiates DMHN from both fixed-landscape CHN and MHN.

\begin{table*}[]
	\centering
	\caption{
		\textbf{Memory capacity evaluation.} Retrieval accuracy versus memory load, showing a clear overall advantage of DMHN over CHN and MHN.
		\label{tab:retrieval-results}
	}
	\vspace{0.2cm}
	\scriptsize
	\setlength{\tabcolsep}{2.7pt}
\begin{tabular}{lccccccccccccccccccccc}
	\toprule
	& \multicolumn{7}{c}{CHN} & \multicolumn{7}{c}{MHN} & \multicolumn{7}{c}{$\mathbf{DMHN}$} \\
	\cmidrule(lr){2-8}
	\cmidrule(lr){9-15}
	\cmidrule(lr){16-22}
	dataset/noise & 0.2N & 0.5N & 1N & 2N & 3N & 5N & 10N & 0.2N & 0.5N & 1N & 2N & 3N & 5N & 10N & 0.2N & 0.5N & 1N & 2N & 3N & 5N & 10N \\
	\midrule
	benchmark/flip & \cellcolor[rgb]{0.99,0.54,0.42}0.80 & \cellcolor[rgb]{1.00,0.90,0.86}0.17 & \cellcolor[rgb]{1.00,0.95,0.93}0.03 & \cellcolor[rgb]{1.00,0.96,0.94}0.00 & \cellcolor[rgb]{1.00,0.96,0.94}0.00 & \cellcolor[rgb]{1.00,0.96,0.94}0.00 & \cellcolor[rgb]{1.00,0.96,0.94}0.00 & \cellcolor[rgb]{0.98,0.41,0.29}1.00 & \cellcolor[rgb]{0.98,0.41,0.29}1.00 & \cellcolor[rgb]{0.99,0.46,0.33}0.93 & \cellcolor[rgb]{1.00,0.88,0.82}0.26 & \cellcolor[rgb]{1.00,0.95,0.92}0.05 & \cellcolor[rgb]{1.00,0.96,0.94}0.00 & \cellcolor[rgb]{1.00,0.96,0.94}0.00 & \cellcolor[rgb]{0.98,0.41,0.29}1.00 & \cellcolor[rgb]{0.98,0.41,0.29}1.00 & \cellcolor[rgb]{0.98,0.43,0.31}0.98 & \cellcolor[rgb]{1.00,0.88,0.83}0.25 & \cellcolor[rgb]{1.00,0.92,0.88}0.13 & \cellcolor[rgb]{1.00,0.93,0.90}0.08 & \cellcolor[rgb]{1.00,0.96,0.94}0.01 \\
	\midrule
	imbalanced (p=0.1)/mask & \cellcolor[rgb]{1.00,0.96,0.94}0.00 & \cellcolor[rgb]{1.00,0.96,0.94}0.00 & \cellcolor[rgb]{1.00,0.96,0.94}0.00 & \cellcolor[rgb]{1.00,0.96,0.94}0.00 & \cellcolor[rgb]{1.00,0.96,0.94}0.00 & \cellcolor[rgb]{1.00,0.96,0.94}0.00 & \cellcolor[rgb]{1.00,0.96,0.94}0.00 & \cellcolor[rgb]{1.00,0.96,0.94}0.00 & \cellcolor[rgb]{1.00,0.96,0.94}0.00 & \cellcolor[rgb]{1.00,0.96,0.94}0.00 & \cellcolor[rgb]{1.00,0.96,0.94}0.00 & \cellcolor[rgb]{1.00,0.96,0.94}0.00 & \cellcolor[rgb]{1.00,0.96,0.94}0.00 & \cellcolor[rgb]{1.00,0.96,0.94}0.00 & \cellcolor[rgb]{0.98,0.42,0.29}1.00 & \cellcolor[rgb]{0.98,0.42,0.29}0.99 & \cellcolor[rgb]{0.99,0.46,0.33}0.93 & \cellcolor[rgb]{0.99,0.61,0.49}0.70 & \cellcolor[rgb]{0.99,0.70,0.59}0.56 & \cellcolor[rgb]{0.99,0.69,0.58}0.56 & \cellcolor[rgb]{0.99,0.67,0.56}0.60 \\
	imbalanced (p=0.2)/mask & \cellcolor[rgb]{1.00,0.96,0.94}0.00 & \cellcolor[rgb]{1.00,0.96,0.94}0.00 & \cellcolor[rgb]{1.00,0.96,0.94}0.00 & \cellcolor[rgb]{1.00,0.96,0.94}0.00 & \cellcolor[rgb]{1.00,0.96,0.94}0.00 & \cellcolor[rgb]{1.00,0.96,0.94}0.00 & \cellcolor[rgb]{1.00,0.96,0.94}0.01 & \cellcolor[rgb]{0.99,0.56,0.43}0.77 & \cellcolor[rgb]{1.00,0.96,0.94}0.00 & \cellcolor[rgb]{1.00,0.96,0.94}0.00 & \cellcolor[rgb]{1.00,0.96,0.94}0.00 & \cellcolor[rgb]{1.00,0.96,0.94}0.00 & \cellcolor[rgb]{1.00,0.96,0.94}0.00 & \cellcolor[rgb]{1.00,0.96,0.94}0.00 & \cellcolor[rgb]{0.98,0.41,0.29}1.00 & \cellcolor[rgb]{0.98,0.42,0.29}1.00 & \cellcolor[rgb]{0.99,0.45,0.33}0.94 & \cellcolor[rgb]{0.99,0.61,0.49}0.70 & \cellcolor[rgb]{0.99,0.73,0.62}0.51 & \cellcolor[rgb]{0.99,0.79,0.71}0.40 & \cellcolor[rgb]{0.99,0.81,0.74}0.36 \\
	imbalanced (p=0.3)/mask & \cellcolor[rgb]{1.00,0.93,0.90}0.08 & \cellcolor[rgb]{1.00,0.96,0.94}0.00 & \cellcolor[rgb]{1.00,0.96,0.94}0.00 & \cellcolor[rgb]{1.00,0.96,0.94}0.00 & \cellcolor[rgb]{1.00,0.96,0.94}0.01 & \cellcolor[rgb]{1.00,0.95,0.93}0.03 & \cellcolor[rgb]{1.00,0.93,0.90}0.09 & \cellcolor[rgb]{0.98,0.41,0.29}1.00 & \cellcolor[rgb]{0.99,0.53,0.40}0.82 & \cellcolor[rgb]{1.00,0.91,0.86}0.17 & \cellcolor[rgb]{1.00,0.96,0.93}0.02 & \cellcolor[rgb]{1.00,0.96,0.94}0.00 & \cellcolor[rgb]{1.00,0.96,0.94}0.00 & \cellcolor[rgb]{1.00,0.96,0.94}0.00 & \cellcolor[rgb]{0.98,0.42,0.29}1.00 & \cellcolor[rgb]{0.98,0.42,0.29}1.00 & \cellcolor[rgb]{0.99,0.46,0.34}0.93 & \cellcolor[rgb]{0.99,0.62,0.50}0.67 & \cellcolor[rgb]{0.99,0.73,0.62}0.51 & \cellcolor[rgb]{0.99,0.81,0.73}0.37 & \cellcolor[rgb]{1.00,0.87,0.82}0.26 \\
	imbalanced (p=0.4)/mask & \cellcolor[rgb]{0.99,0.73,0.62}0.51 & \cellcolor[rgb]{1.00,0.93,0.90}0.09 & \cellcolor[rgb]{1.00,0.95,0.93}0.03 & \cellcolor[rgb]{1.00,0.94,0.91}0.06 & \cellcolor[rgb]{1.00,0.93,0.90}0.08 & \cellcolor[rgb]{1.00,0.92,0.88}0.14 & \cellcolor[rgb]{1.00,0.91,0.87}0.16 & \cellcolor[rgb]{0.98,0.41,0.29}1.00 & \cellcolor[rgb]{0.98,0.41,0.29}1.00 & \cellcolor[rgb]{0.99,0.49,0.36}0.89 & \cellcolor[rgb]{1.00,0.89,0.84}0.21 & \cellcolor[rgb]{1.00,0.94,0.92}0.06 & \cellcolor[rgb]{1.00,0.96,0.94}0.01 & \cellcolor[rgb]{1.00,0.96,0.94}0.00 & \cellcolor[rgb]{0.98,0.42,0.29}1.00 & \cellcolor[rgb]{0.98,0.42,0.29}0.99 & \cellcolor[rgb]{0.99,0.48,0.35}0.90 & \cellcolor[rgb]{0.99,0.65,0.54}0.63 & \cellcolor[rgb]{0.99,0.75,0.66}0.46 & \cellcolor[rgb]{0.99,0.83,0.76}0.34 & \cellcolor[rgb]{1.00,0.87,0.82}0.26 \\
	imbalanced (p=0.5)/mask & \cellcolor[rgb]{0.99,0.55,0.43}0.78 & \cellcolor[rgb]{1.00,0.91,0.87}0.16 & \cellcolor[rgb]{1.00,0.93,0.90}0.09 & \cellcolor[rgb]{1.00,0.92,0.88}0.13 & \cellcolor[rgb]{1.00,0.91,0.86}0.17 & \cellcolor[rgb]{1.00,0.90,0.85}0.20 & \cellcolor[rgb]{1.00,0.90,0.85}0.20 & \cellcolor[rgb]{0.98,0.41,0.29}1.00 & \cellcolor[rgb]{0.98,0.41,0.29}1.00 & \cellcolor[rgb]{0.99,0.46,0.34}0.92 & \cellcolor[rgb]{1.00,0.89,0.83}0.23 & \cellcolor[rgb]{1.00,0.95,0.93}0.04 & \cellcolor[rgb]{1.00,0.96,0.94}0.00 & \cellcolor[rgb]{1.00,0.96,0.94}0.00 & \cellcolor[rgb]{0.98,0.41,0.29}1.00 & \cellcolor[rgb]{0.98,0.42,0.30}0.99 & \cellcolor[rgb]{0.99,0.48,0.35}0.90 & \cellcolor[rgb]{0.99,0.65,0.53}0.63 & \cellcolor[rgb]{0.99,0.75,0.66}0.46 & \cellcolor[rgb]{0.99,0.83,0.76}0.34 & \cellcolor[rgb]{1.00,0.89,0.84}0.23 \\
	\addlinespace
	imbalanced (p=0.1)/gaussian & \cellcolor[rgb]{1.00,0.96,0.94}0.00 & \cellcolor[rgb]{1.00,0.96,0.94}0.00 & \cellcolor[rgb]{1.00,0.96,0.94}0.00 & \cellcolor[rgb]{1.00,0.96,0.94}0.00 & \cellcolor[rgb]{1.00,0.96,0.94}0.00 & \cellcolor[rgb]{1.00,0.96,0.94}0.00 & \cellcolor[rgb]{1.00,0.96,0.94}0.00 & \cellcolor[rgb]{1.00,0.96,0.94}0.00 & \cellcolor[rgb]{1.00,0.96,0.94}0.00 & \cellcolor[rgb]{1.00,0.96,0.94}0.00 & \cellcolor[rgb]{1.00,0.96,0.94}0.00 & \cellcolor[rgb]{1.00,0.96,0.94}0.00 & \cellcolor[rgb]{1.00,0.96,0.94}0.00 & \cellcolor[rgb]{1.00,0.96,0.94}0.00 & \cellcolor[rgb]{0.98,0.41,0.29}1.00 & \cellcolor[rgb]{0.98,0.41,0.29}1.00 & \cellcolor[rgb]{0.98,0.43,0.30}0.98 & \cellcolor[rgb]{0.99,0.56,0.44}0.77 & \cellcolor[rgb]{0.99,0.76,0.67}0.46 & \cellcolor[rgb]{0.99,0.82,0.75}0.34 & \cellcolor[rgb]{0.99,0.86,0.79}0.29 \\
	imbalanced (p=0.2)/gaussian & \cellcolor[rgb]{1.00,0.96,0.94}0.00 & \cellcolor[rgb]{1.00,0.96,0.94}0.00 & \cellcolor[rgb]{1.00,0.96,0.94}0.00 & \cellcolor[rgb]{1.00,0.96,0.94}0.00 & \cellcolor[rgb]{1.00,0.96,0.94}0.00 & \cellcolor[rgb]{1.00,0.96,0.94}0.00 & \cellcolor[rgb]{1.00,0.96,0.94}0.00 & \cellcolor[rgb]{0.99,0.56,0.43}0.78 & \cellcolor[rgb]{1.00,0.96,0.94}0.00 & \cellcolor[rgb]{1.00,0.96,0.94}0.00 & \cellcolor[rgb]{1.00,0.96,0.94}0.00 & \cellcolor[rgb]{1.00,0.96,0.94}0.00 & \cellcolor[rgb]{1.00,0.96,0.94}0.00 & \cellcolor[rgb]{1.00,0.96,0.94}0.00 & \cellcolor[rgb]{0.98,0.41,0.29}1.00 & \cellcolor[rgb]{0.98,0.42,0.29}1.00 & \cellcolor[rgb]{0.98,0.42,0.30}0.99 & \cellcolor[rgb]{0.99,0.57,0.44}0.76 & \cellcolor[rgb]{0.99,0.76,0.67}0.46 & \cellcolor[rgb]{1.00,0.90,0.86}0.18 & \cellcolor[rgb]{1.00,0.92,0.88}0.13 \\
	imbalanced (p=0.3)/gaussian & \cellcolor[rgb]{1.00,0.95,0.93}0.03 & \cellcolor[rgb]{1.00,0.96,0.94}0.00 & \cellcolor[rgb]{1.00,0.96,0.94}0.00 & \cellcolor[rgb]{1.00,0.96,0.94}0.00 & \cellcolor[rgb]{1.00,0.96,0.94}0.00 & \cellcolor[rgb]{1.00,0.96,0.94}0.00 & \cellcolor[rgb]{1.00,0.96,0.94}0.00 & \cellcolor[rgb]{0.98,0.41,0.29}1.00 & \cellcolor[rgb]{0.99,0.53,0.40}0.82 & \cellcolor[rgb]{1.00,0.91,0.86}0.17 & \cellcolor[rgb]{1.00,0.96,0.93}0.02 & \cellcolor[rgb]{1.00,0.96,0.94}0.00 & \cellcolor[rgb]{1.00,0.96,0.94}0.00 & \cellcolor[rgb]{1.00,0.96,0.94}0.00 & \cellcolor[rgb]{0.98,0.41,0.29}1.00 & \cellcolor[rgb]{0.98,0.41,0.29}1.00 & \cellcolor[rgb]{0.98,0.42,0.30}0.99 & \cellcolor[rgb]{0.99,0.58,0.46}0.74 & \cellcolor[rgb]{0.99,0.79,0.70}0.41 & \cellcolor[rgb]{1.00,0.89,0.85}0.21 & \cellcolor[rgb]{1.00,0.93,0.90}0.09 \\
	imbalanced (p=0.4)/gaussian & \cellcolor[rgb]{0.99,0.76,0.67}0.46 & \cellcolor[rgb]{1.00,0.95,0.93}0.02 & \cellcolor[rgb]{1.00,0.96,0.94}0.00 & \cellcolor[rgb]{1.00,0.96,0.94}0.00 & \cellcolor[rgb]{1.00,0.96,0.94}0.00 & \cellcolor[rgb]{1.00,0.96,0.94}0.00 & \cellcolor[rgb]{1.00,0.96,0.94}0.01 & \cellcolor[rgb]{0.98,0.41,0.29}1.00 & \cellcolor[rgb]{0.98,0.41,0.29}1.00 & \cellcolor[rgb]{0.99,0.49,0.36}0.89 & \cellcolor[rgb]{1.00,0.89,0.84}0.21 & \cellcolor[rgb]{1.00,0.94,0.92}0.06 & \cellcolor[rgb]{1.00,0.96,0.94}0.01 & \cellcolor[rgb]{1.00,0.96,0.94}0.00 & \cellcolor[rgb]{0.98,0.41,0.29}1.00 & \cellcolor[rgb]{0.98,0.41,0.29}1.00 & \cellcolor[rgb]{0.98,0.42,0.30}0.99 & \cellcolor[rgb]{0.99,0.62,0.50}0.68 & \cellcolor[rgb]{0.99,0.82,0.74}0.36 & \cellcolor[rgb]{1.00,0.90,0.86}0.18 & \cellcolor[rgb]{1.00,0.93,0.90}0.09 \\
	imbalanced (p=0.5)/gaussian & \cellcolor[rgb]{0.99,0.54,0.41}0.81 & \cellcolor[rgb]{1.00,0.93,0.90}0.09 & \cellcolor[rgb]{1.00,0.95,0.93}0.03 & \cellcolor[rgb]{1.00,0.95,0.93}0.03 & \cellcolor[rgb]{1.00,0.96,0.93}0.02 & \cellcolor[rgb]{1.00,0.95,0.93}0.03 & \cellcolor[rgb]{1.00,0.95,0.93}0.04 & \cellcolor[rgb]{0.98,0.41,0.29}1.00 & \cellcolor[rgb]{0.98,0.41,0.29}1.00 & \cellcolor[rgb]{0.99,0.46,0.34}0.92 & \cellcolor[rgb]{1.00,0.89,0.83}0.23 & \cellcolor[rgb]{1.00,0.95,0.92}0.04 & \cellcolor[rgb]{1.00,0.96,0.94}0.00 & \cellcolor[rgb]{1.00,0.96,0.94}0.00 & \cellcolor[rgb]{0.98,0.41,0.29}1.00 & \cellcolor[rgb]{0.98,0.41,0.29}1.00 & \cellcolor[rgb]{0.98,0.42,0.30}0.99 & \cellcolor[rgb]{0.99,0.61,0.49}0.69 & \cellcolor[rgb]{0.99,0.81,0.73}0.38 & \cellcolor[rgb]{1.00,0.91,0.86}0.17 & \cellcolor[rgb]{1.00,0.93,0.90}0.08 \\
	\addlinespace
	average & \cellcolor[rgb]{1.00,0.87,0.81}0.27 & \cellcolor[rgb]{1.00,0.95,0.93}0.04 & \cellcolor[rgb]{1.00,0.96,0.93}0.02 & \cellcolor[rgb]{1.00,0.96,0.93}0.02 & \cellcolor[rgb]{1.00,0.95,0.93}0.03 & \cellcolor[rgb]{1.00,0.95,0.92}0.04 & \cellcolor[rgb]{1.00,0.95,0.92}0.05 & \cellcolor[rgb]{0.99,0.57,0.45}0.75 & \cellcolor[rgb]{0.99,0.69,0.58}0.56 & \cellcolor[rgb]{0.99,0.80,0.71}0.40 & \cellcolor[rgb]{1.00,0.93,0.90}0.09 & \cellcolor[rgb]{1.00,0.96,0.93}0.02 & \cellcolor[rgb]{1.00,0.96,0.94}0.00 & \cellcolor[rgb]{1.00,0.96,0.94}0.00 & \cellcolor[rgb]{0.98,0.42,0.29}1.00 & \cellcolor[rgb]{0.98,0.42,0.29}1.00 & \cellcolor[rgb]{0.98,0.44,0.32}0.95 & \cellcolor[rgb]{0.99,0.61,0.49}0.70 & \cellcolor[rgb]{0.99,0.76,0.67}0.46 & \cellcolor[rgb]{0.99,0.85,0.78}0.31 & \cellcolor[rgb]{1.00,0.88,0.83}0.24 \\
	\midrule
	MNIST/mask & \cellcolor[rgb]{1.00,0.96,0.94}0.00 & \cellcolor[rgb]{1.00,0.96,0.94}0.00 & \cellcolor[rgb]{1.00,0.96,0.94}0.00 & \cellcolor[rgb]{1.00,0.96,0.94}0.00 & \cellcolor[rgb]{1.00,0.96,0.94}0.00 & \cellcolor[rgb]{1.00,0.96,0.94}0.00 & \cellcolor[rgb]{1.00,0.96,0.94}0.00 & \cellcolor[rgb]{1.00,0.96,0.94}0.00 & \cellcolor[rgb]{1.00,0.96,0.94}0.00 & \cellcolor[rgb]{1.00,0.96,0.94}0.00 & \cellcolor[rgb]{1.00,0.96,0.94}0.00 & \cellcolor[rgb]{1.00,0.96,0.94}0.00 & \cellcolor[rgb]{1.00,0.96,0.94}0.00 & \cellcolor[rgb]{1.00,0.96,0.94}0.00 & \cellcolor[rgb]{0.99,0.45,0.33}0.94 & \cellcolor[rgb]{0.99,0.51,0.39}0.85 & \cellcolor[rgb]{0.99,0.56,0.44}0.77 & \cellcolor[rgb]{0.99,0.60,0.48}0.71 & \cellcolor[rgb]{0.99,0.62,0.50}0.68 & \cellcolor[rgb]{0.99,0.63,0.51}0.66 & \cellcolor[rgb]{0.99,0.65,0.54}0.63 \\
	MNIST/gaussian & \cellcolor[rgb]{1.00,0.96,0.94}0.00 & \cellcolor[rgb]{1.00,0.96,0.94}0.00 & \cellcolor[rgb]{1.00,0.96,0.94}0.00 & \cellcolor[rgb]{1.00,0.96,0.94}0.00 & \cellcolor[rgb]{1.00,0.96,0.94}0.00 & \cellcolor[rgb]{1.00,0.96,0.94}0.00 & \cellcolor[rgb]{1.00,0.96,0.94}0.00 & \cellcolor[rgb]{1.00,0.96,0.94}0.00 & \cellcolor[rgb]{1.00,0.96,0.94}0.00 & \cellcolor[rgb]{1.00,0.96,0.94}0.00 & \cellcolor[rgb]{1.00,0.96,0.94}0.00 & \cellcolor[rgb]{1.00,0.96,0.94}0.00 & \cellcolor[rgb]{1.00,0.96,0.94}0.00 & \cellcolor[rgb]{1.00,0.96,0.94}0.00 & \cellcolor[rgb]{0.98,0.42,0.29}1.00 & \cellcolor[rgb]{0.98,0.43,0.30}0.98 & \cellcolor[rgb]{0.99,0.47,0.35}0.91 & \cellcolor[rgb]{0.99,0.53,0.40}0.82 & \cellcolor[rgb]{0.99,0.56,0.43}0.78 & \cellcolor[rgb]{0.99,0.58,0.46}0.73 & \cellcolor[rgb]{0.99,0.62,0.50}0.68 \\
	\addlinespace
	average & \cellcolor[rgb]{1.00,0.96,0.94}0.00 & \cellcolor[rgb]{1.00,0.96,0.94}0.00 & \cellcolor[rgb]{1.00,0.96,0.94}0.00 & \cellcolor[rgb]{1.00,0.96,0.94}0.00 & \cellcolor[rgb]{1.00,0.96,0.94}0.00 & \cellcolor[rgb]{1.00,0.96,0.94}0.00 & \cellcolor[rgb]{1.00,0.96,0.94}0.00 & \cellcolor[rgb]{1.00,0.96,0.94}0.00 & \cellcolor[rgb]{1.00,0.96,0.94}0.00 & \cellcolor[rgb]{1.00,0.96,0.94}0.00 & \cellcolor[rgb]{1.00,0.96,0.94}0.00 & \cellcolor[rgb]{1.00,0.96,0.94}0.00 & \cellcolor[rgb]{1.00,0.96,0.94}0.00 & \cellcolor[rgb]{1.00,0.96,0.94}0.00 & \cellcolor[rgb]{0.98,0.43,0.31}0.97 & \cellcolor[rgb]{0.99,0.47,0.34}0.92 & \cellcolor[rgb]{0.99,0.52,0.39}0.84 & \cellcolor[rgb]{0.99,0.56,0.44}0.77 & \cellcolor[rgb]{0.99,0.59,0.46}0.73 & \cellcolor[rgb]{0.99,0.61,0.49}0.70 & \cellcolor[rgb]{0.99,0.64,0.52}0.66 \\
	\midrule
	CIFAR10/mask & \cellcolor[rgb]{1.00,0.96,0.94}0.00 & \cellcolor[rgb]{1.00,0.96,0.94}0.00 & \cellcolor[rgb]{1.00,0.96,0.94}0.00 & \cellcolor[rgb]{1.00,0.96,0.94}0.00 & \cellcolor[rgb]{1.00,0.96,0.94}0.00 & \cellcolor[rgb]{1.00,0.96,0.94}0.00 & \cellcolor[rgb]{1.00,0.96,0.94}0.00 & \cellcolor[rgb]{0.98,0.42,0.29}1.00 & \cellcolor[rgb]{0.99,0.46,0.34}0.93 & \cellcolor[rgb]{0.99,0.77,0.68}0.44 & \cellcolor[rgb]{1.00,0.90,0.86}0.18 & \cellcolor[rgb]{1.00,0.91,0.87}0.15 & \cellcolor[rgb]{1.00,0.91,0.87}0.15 & \cellcolor[rgb]{1.00,0.91,0.87}0.15 & \cellcolor[rgb]{0.98,0.42,0.29}0.99 & \cellcolor[rgb]{0.99,0.54,0.41}0.81 & \cellcolor[rgb]{0.99,0.56,0.44}0.77 & \cellcolor[rgb]{0.99,0.59,0.46}0.73 & \cellcolor[rgb]{0.99,0.60,0.48}0.71 & \cellcolor[rgb]{0.99,0.59,0.46}0.73 & \cellcolor[rgb]{0.99,0.59,0.47}0.72 \\
	CIFAR10/gaussian & \cellcolor[rgb]{1.00,0.96,0.94}0.00 & \cellcolor[rgb]{1.00,0.96,0.94}0.00 & \cellcolor[rgb]{1.00,0.96,0.94}0.00 & \cellcolor[rgb]{1.00,0.96,0.94}0.00 & \cellcolor[rgb]{1.00,0.96,0.94}0.00 & \cellcolor[rgb]{1.00,0.96,0.94}0.00 & \cellcolor[rgb]{1.00,0.96,0.94}0.00 & \cellcolor[rgb]{0.98,0.41,0.29}1.00 & \cellcolor[rgb]{0.99,0.46,0.34}0.93 & \cellcolor[rgb]{0.99,0.77,0.68}0.44 & \cellcolor[rgb]{1.00,0.90,0.85}0.19 & \cellcolor[rgb]{1.00,0.91,0.86}0.17 & \cellcolor[rgb]{1.00,0.91,0.86}0.17 & \cellcolor[rgb]{1.00,0.90,0.86}0.18 & \cellcolor[rgb]{0.98,0.42,0.29}1.00 & \cellcolor[rgb]{0.98,0.42,0.30}0.99 & \cellcolor[rgb]{0.98,0.42,0.29}0.99 & \cellcolor[rgb]{0.98,0.44,0.32}0.95 & \cellcolor[rgb]{0.99,0.49,0.36}0.88 & \cellcolor[rgb]{0.99,0.54,0.42}0.80 & \cellcolor[rgb]{0.99,0.55,0.43}0.79 \\
	\addlinespace
	average & \cellcolor[rgb]{1.00,0.96,0.94}0.00 & \cellcolor[rgb]{1.00,0.96,0.94}0.00 & \cellcolor[rgb]{1.00,0.96,0.94}0.00 & \cellcolor[rgb]{1.00,0.96,0.94}0.00 & \cellcolor[rgb]{1.00,0.96,0.94}0.00 & \cellcolor[rgb]{1.00,0.96,0.94}0.00 & \cellcolor[rgb]{1.00,0.96,0.94}0.00 & \cellcolor[rgb]{0.98,0.42,0.29}1.00 & \cellcolor[rgb]{0.99,0.46,0.34}0.93 & \cellcolor[rgb]{0.99,0.77,0.68}0.44 & \cellcolor[rgb]{1.00,0.90,0.86}0.19 & \cellcolor[rgb]{1.00,0.91,0.87}0.16 & \cellcolor[rgb]{1.00,0.91,0.87}0.16 & \cellcolor[rgb]{1.00,0.91,0.86}0.17 & \cellcolor[rgb]{0.98,0.42,0.29}1.00 & \cellcolor[rgb]{0.99,0.48,0.35}0.90 & \cellcolor[rgb]{0.99,0.49,0.37}0.88 & \cellcolor[rgb]{0.99,0.52,0.39}0.84 & \cellcolor[rgb]{0.99,0.54,0.42}0.80 & \cellcolor[rgb]{0.99,0.56,0.44}0.77 & \cellcolor[rgb]{0.99,0.57,0.45}0.75 \\
	\midrule
	average overall & \cellcolor[rgb]{1.00,0.87,0.81}0.27 & \cellcolor[rgb]{1.00,0.95,0.92}0.05 & \cellcolor[rgb]{1.00,0.96,0.94}0.01 & \cellcolor[rgb]{1.00,0.96,0.94}0.01 & \cellcolor[rgb]{1.00,0.96,0.94}0.01 & \cellcolor[rgb]{1.00,0.96,0.94}0.01 & \cellcolor[rgb]{1.00,0.96,0.94}0.01 & \cellcolor[rgb]{0.99,0.61,0.49}0.69 & \cellcolor[rgb]{0.99,0.66,0.54}0.62 & \cellcolor[rgb]{0.99,0.77,0.68}0.44 & \cellcolor[rgb]{1.00,0.92,0.88}0.13 & \cellcolor[rgb]{1.00,0.94,0.92}0.06 & \cellcolor[rgb]{1.00,0.95,0.92}0.04 & \cellcolor[rgb]{1.00,0.95,0.92}0.04 & \cellcolor[rgb]{0.98,0.42,0.30}0.99 & \cellcolor[rgb]{0.98,0.44,0.32}0.95 & \cellcolor[rgb]{0.99,0.47,0.35}0.91 & \cellcolor[rgb]{0.99,0.65,0.53}0.64 & \cellcolor[rgb]{0.99,0.72,0.61}0.53 & \cellcolor[rgb]{0.99,0.75,0.66}0.46 & \cellcolor[rgb]{0.99,0.78,0.70}0.42 \\
	\bottomrule
\end{tabular}

\end{table*}

\subsection{Dynamic manifold reorganization underlies DMHN's dynamics}
We tested whether memory retrieval in DMHN is governed by a fixed attractor landscape with cue-dependent initial conditions or by cue-dependent reorganization of the underlying energy landscape.
In CHN and Static-DMHN (S-DMHN),
retrieval unfolds on a fixed energy landscape, such that variations in the cue lead to convergence toward different pre-existing attractor basins.
By contrast, in DMHN, retrieval trajectories are shaped by cue-conditioned modulation of the energy landscape, giving rise to attractor manifolds whose geometry deforms systematically with the cue (Fig.~\ref{fig:diagram}(e--h)).

Consistent with this observation, as the cue is continuously varied, DMHN trajectories evolve smoothly along dynamic manifolds induced by a continuously deforming energy landscape, rather than transitioning between static attractor basins (Fig.~\ref{fig:overlap}, SI Appendix, Movies S1--S4).
Although each retrieval episode converges to a stable fixed point, the location of the fixed point and the local geometry of its surrounding attractor basin vary continuously with the cue, indicating that the effective energy landscape is reorganized across contexts.
This behavior differs fundamentally from CHN and S-DMHN, in which retrieval dynamics remain confined to a fixed attractor geometry.
These findings indicate that associative retrieval in DMHN is organized by cue-conditioned manifold reorganization, consistent with the presence of a dynamically modulated energy landscape during memory recall (Fig.~\ref{fig:diagram}(h)).

\subsection{High-capacity and robust retrieval emerges from dynamic manifold reorganization}
\begin{figure*}[t]
	\centering
	\includegraphics[width=\linewidth]{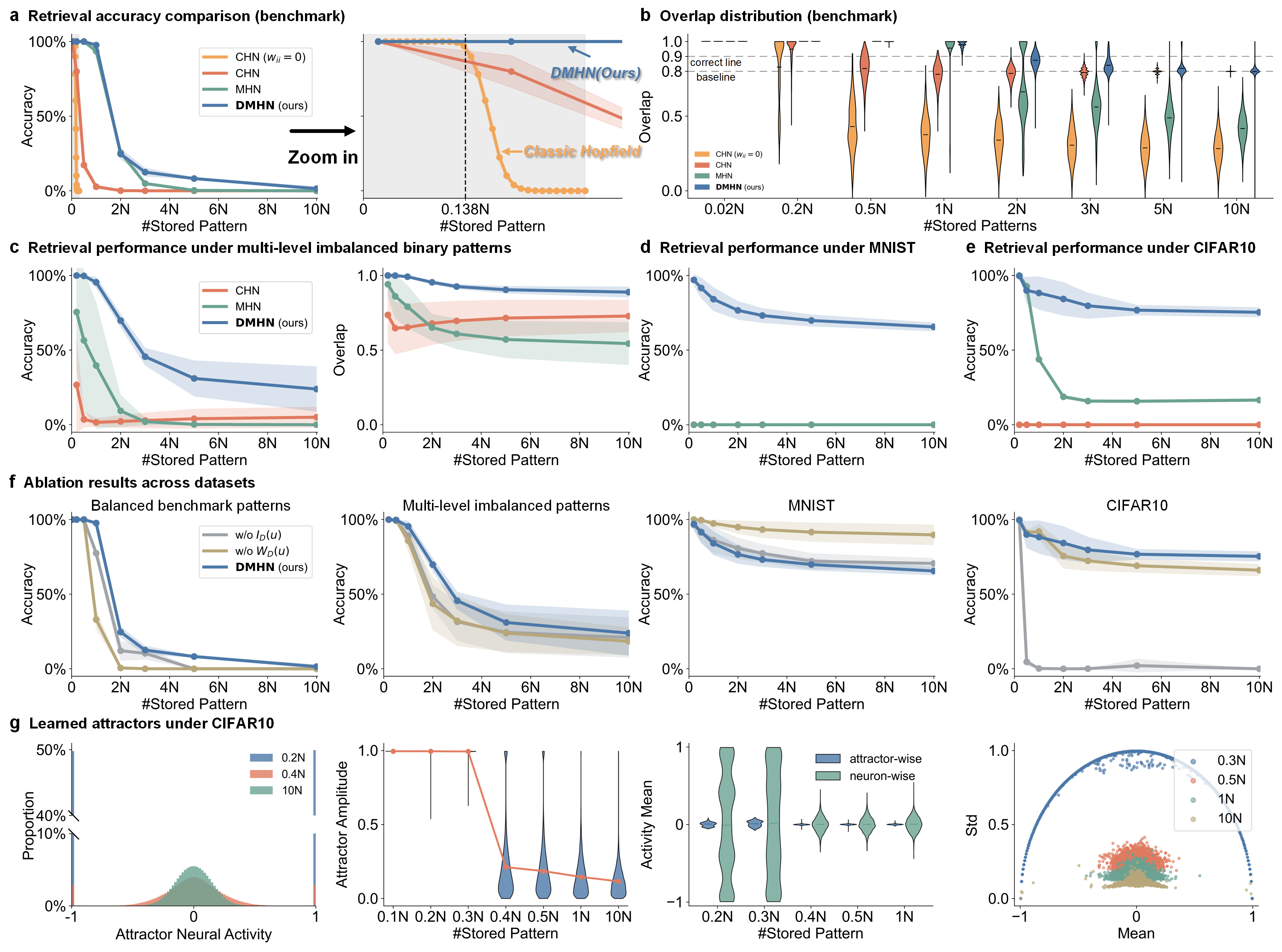}
	\caption{
		\textbf{Pattern retrieval performance and robustness across datasets.}
		(a) Retrieval accuracy as a function of stored patterns for CHN, MHN, and DMHN. Right: zoomed view near the classical capacity limit.
		(b) Overlap distributions between retrieved states and target patterns across memory loads. Dashed lines denote the correct-retrieval threshold and the baseline overlap of the initial state.
		(c--e) Retrieval performance under heterogeneous binary patterns (imbalanced, MNIST) and continuous-valued images (CIFAR10).
		(f) Ablation results across datasets.
		(g) Attractor activity statistics learned under CIFAR10 evaluation.
	}
	\label{fig:pattern-results}
\end{figure*}

Retrieval reliability under high memory load, where interference among stored patterns becomes prominent, differs markedly across models.
In CHN and MHN, interference among attractors leads to spurious fixed points and degraded retrieval performance as the number of stored memories increases.
Accordingly, retrieval accuracy in CHN and MHN decreased sharply with increasing memory load across all tested noise models and datasets.
By contrast, DMHN maintained reliable retrieval across a substantially wider range of loads (Fig.~\ref{fig:pattern-results}(a--e), Tab.~\ref{tab:retrieval-results}).
For example, when storing $2N$ patterns, DMHN achieved an average accuracy of 64\%, compared with 1\% and 13\% for CHN and MHN, respectively (Tab.~\ref{tab:retrieval-results}).

Retrieval trajectories under high memory load further revealed distinct dynamical regimes across models (Fig.~\ref{fig:retrieval}(d)).
Under $2N$ stored patterns, CHN and MHN exhibited densely overlapping attractors and frequent convergence to incorrect or mixed states.
In contrast, DMHN trajectories converged toward the target memory along attractor geometries reshaped by the cue, consistent with dynamic reorganization of the energy landscape that mitigates attractor interference.
Collectively, these observations suggest that cue-conditioned manifold reorganization mitigates attractor interference under increasing memory load, thereby supporting high-capacity and robust retrieval in DMHN.

\subsection{Robust associative retrieval across heterogeneous and structured memory statistics}
Associative memories in biological and real-world settings are characterized by heterogeneous activation statistics and structured correlations rather than balanced random patterns, differing substantially from the assumptions underlying classical associative memory models \cite{hebbian, hopfield1984neurons}.

Retrieval performance under heterogeneous activation statistics was assessed on random binary patterns with systematically varied activation ratios ($p=\frac{\#1}{\#1+\#-1}$, corresponding to sparsity levels ranging from 0.1 to 0.5).
Across this wide range of imbalance levels, DMHN consistently achieved higher retrieval accuracy than CHN and MHN under both masking and Gaussian corruption (Fig.~\ref{fig:retrieval}(c,d), Fig.~\ref{fig:pattern-results}(c), Tab.~\ref{tab:retrieval-results}, SI Appendix, Figs.~S2--S6).
Whereas CHN and MHN exhibited sharp performance degradation as patterns became increasingly imbalanced (lower $p$), DMHN maintained substantially higher retrieval accuracy across imbalance regimes.

Retrieval on structured natural binary patterns derived from the MNIST dataset revealed a similar trend.
MNIST digits exhibit strong spatial correlations and heterogeneous activation statistics, posing a substantially more challenging setting for fixed-landscape associative memories.
In this regime, CHN and MHN frequently converged to distorted or mixed states corresponding to spurious attractors induced by shared structural features, whereas DMHN reliably reconstructed the target digit under severe corruption (Fig.~\ref{fig:retrieval}(c,d), Fig.~\ref{fig:pattern-results}(d), Tab.~\ref{tab:retrieval-results}, SI Appendix, Fig.~S7).

This robustness further extended to associative retrieval of continuous-valued memories.
When evaluated on continuous image representations derived from CIFAR10, CHN and MHN frequently failed to converge or collapsed to degraded representations, whereas DMHN reliably converged to the target memory while preserving meaningful structure in the retrieved output (Fig.~\ref{fig:retrieval}(c,d), Fig.~\ref{fig:pattern-results}(e), Tab.~\ref{tab:retrieval-results}, SI Appendix, Fig.~S8).
Retrieval is governed by recurrent Hopfield dynamics operating in a continuous latent space, with feedforward encoder-decoder used only for representation mapping (Fig.~\ref{fig:retrieval}(b)).

Across heterogeneous activation statistics, structured correlations, and continuous representations, retrieval in DMHN remains stable in imbalanced and highly correlated regimes where CHN and MHN degrade, consistent with its capacity for dynamic manifold reorganization.

\subsection{Functional dissociation of cue-dependent energy in retrieval dynamics}

We performed systematic ablation studies to assess the contributions of cue-dependent components in DMHN to associative retrieval performance.
Specifically, the dynamic weight term ($W_D$) or the dynamic bias term ($I_D$) was selectively ablated while keeping all other architectural choices and training procedures fixed.

Ablation effects differed systematically across statistical regimes of the stored patterns.
In balanced pattern settings, removal of the $W_D$ term produced a pronounced reduction in retrieval accuracy, whereas ablation of $I_D$ led to a more modest performance decrease (Fig.~\ref{fig:pattern-results}(f;left), SI Appendix, Tabs.~S1,S2).
In contrast, under imbalanced memory distributions, ablation of the $I_D$ term consistently caused a substantial accuracy drop, while removal of $W_D$ had a weaker effect and, in some conditions,	resulted in comparable or marginally higher accuracy (Fig.~\ref{fig:pattern-results}(f;right), SI Appendix, Tabs.~S1,S2).

Retrieval performance exhibits complementary sensitivities, where $W_D$ plays a dominant role in balanced regimes, and $I_D$ becomes critical under highly imbalanced statistics.

\subsection{Attractor activity shifts from polarized to Gaussian as memory load increases}

We analyzed the activity distributions of learned attractors in the CIFAR10 setting as memory load increased (Fig.~\ref{fig:pattern-results}(g)).

With increasing load, attractor activity shifted from strongly polarized values toward centralized distributions.
At low load, activity was concentrated near $\pm1$, whereas higher loads were associated with reduced polarization and smaller activity magnitudes.
The temporal evolution of these distributions during training further reveals a progressive desaturation of attractor activity under higher memory loads (SI Appendix, Movie S5).

Neuron-wise activation statistics remained broadly distributed across memories.
These observations indicate a redistribution of activity under higher memory load, rather than a uniform suppression of neural activity.

\section{Discussion}

We propose Dynamic Manifold Hopfield Networks (DMHN), a dynamical framework for associative memory in which contextual modulation of the energy landscape reorganizes attractor manifolds, enabling cue-conditioned retrieval with enhanced capacity and robustness while preserving continuous attractor dynamics. More broadly, these results support a general mechanistic principle in which flexible yet stable computation arises from dynamic reorganization of underlying manifolds, rather than from static representations or explicit enumeration of attractors or contexts, in both biological and artificial systems.

\paragraph*{Context-dependent manifolds bridge flexibility and stability in associative memory}

A growing body of experimental work suggests that neural population activity evolves on low-dimensional manifolds whose geometry depends on task demands, context, or internal state \cite{gallego2017neural, stringer2019highdimensional}.
Context-dependent remapping in hippocampal and cortical circuits further shows that the same neural population can support distinct representations across conditions, without large-scale changes in connectivity \cite{muller1987effects, wills2005attractor, Bicanski2018, Sanders2020}.

Recent studies suggest that such remapping is usually continuous, with population trajectories deforming smoothly as context varies \cite{remington2018flexible, perich2025neural}.
However, most existing models remain largely descriptive, leaving unresolved how dynamic manifold reorganization is realized mechanistically within a unified dynamical system.

DMHN provide a minimal dynamical instantiation of this idea.
Instead of introducing attractor- or context-specific parameterizations, contextual signals directly modulate the energy landscape, inducing continuous reorganization of attractor manifolds within a single recurrent system.
Retrieval remains governed by continuous attractor dynamics, consistent with theoretical accounts emphasizing smooth flows and low-dimensional structure in neural state space \cite{cunningham2014dimensionality, khona2022attractor}.
Although not intended as a model of a specific circuit, DMHN offer a concrete hypothesis for how context-dependent manifold geometry may emerge from modulated recurrent dynamics.

\paragraph*{Dynamic reorganization of attractor geometry as a principle for flexible memory}

Associative memory models have long been constrained by the geometry of fixed attractor landscapes.
In classical Hopfield networks and related continuous attractor systems, all stored memories must coexist within a single energy landscape, such that increasing memory load inevitably leads to attractor interference, spurious fixed points, and degraded retrieval \cite{hopfield1984neurons, HopfieldCapacity}.
Even approaches that improve storage capacity typically retain a fixed retrieval manifold, redistributing interference rather than resolving it at the level of attractor reorganization \cite{HopfieldCapacityNew, ramsauer2020hopfield, InformationCapacity, krotov2016dense}.

DMHN relax this constraint by allowing the energy landscape to depend on the retrieval cue.
Retrieval dynamics therefore unfold on a family of cue-conditioned manifolds rather than a single static geometry, reducing interference among memories.
From a dynamical systems perspective, this shifts the primary limitation of associative memory from the number of stable attractors to the flexibility with which attractor manifolds can be dynamically reorganized.

Related principles have been explored in studies of task-modulated recurrent dynamics, where contextual signals reshape the effective flow field without altering the underlying network architecture \cite{sussillo2015neural, mastrogiuseppe2018linking}.
DMHN extend this idea to associative memory, showing that dynamic manifold reorganization can substantially improve retrieval under high load and heterogeneous memory statistics.

\paragraph*{Reorganization of attractor statistics under increasing memory load}
In continuous-valued memory settings, we observed systematic changes in attractor activity statistics as memory load increased.
Attractor representations became less polarized and more centrally distributed, while neuron-wise activation remained broadly heterogeneous across memories, consistent with a redistribution of activity rather than uniform suppression.

Similar trade-offs between representational amplitude and population-level diversity have been noted in studies of distributed coding in recurrent networks \cite{babadi2014sparseness}.
Within DMHN, these observations suggest that dynamic manifold reorganization may stabilize retrieval under high load by adjusting representational geometry in continuous attractor systems.

\paragraph*{Dynamic manifolds as a framework for understanding associative memory}
Prior work on associative memory spans several paradigms, including continuous attractor networks defined by fixed energy landscapes, extensions that introduce context-dependent modulation through saliency reweighting or gating mechanisms, and modern attention-based memory models with many-body interactions \cite{hopfield1984neurons, krotov2016dense, ramsauer2020hopfield,betteti2023input,podlaski2025high,kozachkov2025neuronastrocyte}.
While these approaches differ in capacity and learning mechanisms, most either retain a static retrieval geometry or rely on explicit, enumerative parameterization to accommodate multiple contexts or memories.

DMHN address a distinct and previously unresolved challenge within this landscape: how flexible, context-dependent reorganization of associative memory can be achieved within a single continuous dynamical system governed by pairwise synaptic interactions.
By allowing contextual signals to modulate the energy landscape itself, DMHN reshape retrieval manifolds intrinsically, rather than selecting among pre-specified attractor structures.
This dynamic-manifold perspective emphasizes geometric flexibility over static capacity and connects naturally to theoretical frameworks of continuous attractors and population-level neural dynamics \cite{Burak2009, Wimmer2014}.

\paragraph*{Limitations and future directions}
How dynamic manifold reconfiguration can be incorporated into embodied agents operating in interactive environments remains an open question.
In addition, continuous manifold reorganization entails smooth transitions among context-dependent attractor states, raising questions about how such dynamics are stabilized and regulated in biological systems.

Future work may extend DMHN to temporally structured memories, sequential recall, and context-dependent transitions between attractor manifolds.
Exploring biologically plausible mechanisms for contextual modulation, as well as integration with large-scale learned representations, could further clarify dynamic manifolds as a general computational principle for flexible memory in both brains and artificial systems.


\begin{table*}[t]
	\centering
	\caption{
		\textbf{Datasets, corruption models, evaluation metrics, and retrieval criteria.}
		\label{tab:datasets-metrics}
	}
	\vspace{0.2cm}
	\scriptsize
	\setlength{\tabcolsep}{6pt}
	\begin{tabular}{lccc}
		\toprule
		\textbf{Dataset} & \textbf{Corruption} & \textbf{Metric} & \textbf{Retrieval criterion} \\
		\midrule
		Balanced benchmark patterns
		& Random bit flips 
		& Bit-wise error rate ($\mathrm{err}$), overlap ($m$)
		& $\mathrm{err} < 5\%$ \\
		
		Multi-level imbalanced binary patterns
		& Masking / Gaussian noise 
		& Bit-wise error rate ($\mathrm{err}$), overlap ($m$)
		& $\mathrm{err} < 1.5\%$ \\
		
		Structured binary patterns (MNIST)
		& Masking / Gaussian noise 
		& Bit-wise error rate ($\mathrm{err}$)
		& $\mathrm{err} < 1.5\%$ \\
		Continuous-valued image patterns (CIFAR10)
		& Patch masking / Gaussian noise 
		& $\mathrm{MSE}$
		& $\mathrm{MSE} < 25/(HWC)$ \\
		\bottomrule
	\end{tabular}
\end{table*}

\section{Materials and Methods}

We compare Dynamic Manifold Hopfield Networks (DMHN) with Classical Hopfield Networks (CHN) \cite{hopfield1984neurons} and Modern Hopfield Networks (MHN) \cite{ramsauer2020hopfield} under a unified associative retrieval framework.
Models are evaluated on synthetic and natural memory tasks probing retrieval dynamics, capacity, and robustness.
CHN implement fixed attractor dynamics learned by Hebbian rules, MHN provide a high-capacity baseline trained via backpropagation, and DMHN introduce cue-dependent modulation of continuous recurrent dynamics.
Retrieval performance is assessed using matched training protocols and standardized metrics across models.

\subsection*{Classical Hopfield Networks}

Classical Hopfield Networks (CHN) \cite{hopfield1984neurons} are defined over a network state $\mathbf{x}\in\mathbb{R}^N$ with dynamics
\begin{equation}
	\dot{\mathbf{x}} = -\tau \mathbf{x} + \mathbf{W}\,\Phi(\mathbf{x}) + \mathbf{I},
\end{equation}
where $\tau$ is a fixed leak coefficient, $\mathbf{W}$ is a symmetric synaptic weight matrix learned using a variant of the Hebbian rule \cite{HopfieldCapacityNew}, $\Phi(\cdot)$ is a monotonic element-wise nonlinearity, and $\mathbf{I}$ is a bias term.
Throughout this work, $\Phi(\cdot)$ is chosen as $\tanh(\cdot)$.

Under these dynamics, the system admits an energy function
\begin{align}
	E(\mathbf{x})
	=
	&-\frac{1}{2}\Phi(\mathbf{x})^{\top}
	\mathbf{W}
	\Phi(\mathbf{x})
	-\Phi(\mathbf{x})^{\top}
	\mathbf{I}
	+
	\tau
	\int_{0}^{\mathbf{x}} \mathbf{v}\,\Phi'(\mathbf{v})\,\mathrm{d}\mathbf{v},
\end{align}
and converges to stable fixed points corresponding to stored memory patterns.
The retrieval cue specifies the initial condition $\mathbf{x}(0)$, and the synaptic matrix $\mathbf{W}$ is the only learnable component.

\subsection*{Modern Hopfield Networks}

Modern Hopfield Networks (MHN) extend associative memory models by introducing dense, many-body interactions that substantially increase storage capacity \cite{ramsauer2020hopfield}.
In contrast to CHN, memories in MHN can be encoded implicitly in model parameters learned via backpropagation rather than stored explicitly as individual patterns.

Retrieval in MHN is typically formulated as an update rule of the form
\begin{equation}
	\mathbf{x}_{t+1}
	=
	\sum_{\mu=1}^{M}
	\mathbf{y}^{\mu}\,
	\frac{\exp\!\big(\beta \langle \mathbf{x}_t, \mathbf{k}^{\mu} \rangle\big)}
	{\sum_{\nu=1}^{M}\exp\!\big(\beta \langle \mathbf{x}_t, \mathbf{k}^{\nu} \rangle\big)}.
\end{equation}
where $\{\mathbf{k}^{\mu}, \mathbf{y}^{\mu}\}$ denote learned key-value representations and $\beta$ controls the sharpness of the update.
These parameters are optimized using gradient-based training in our experiments.

Unlike CHN and DMHN, MHN do not define continuous-time recurrent dynamics governed by neuron-to-neuron synaptic interactions.
Instead, retrieval proceeds through attention-like many-body interactions.
In this work, MHN serve as a high-capacity associative memory baseline that lacks classical continuous attractor dynamics.

\subsection*{Dynamic Manifold Hopfield Networks}

Dynamic Manifold Hopfield Networks (DMHN) extend CHN by allowing retrieval cues to modulate network dynamics, rather than acting solely as initial states.
This cue-dependent modulation enables context-dependent reorganization of the energy landscape, alleviating attractor interference that arises when all memories must be embedded within a single fixed geometry.
Crucially, this flexibility is achieved while preserving pairwise recurrent interactions and continuous attractor dynamics within a single integrated system.

In DMHN, synaptic and bias terms are decomposed into cue-independent (static) and cue-dependent (dynamic) components, yielding the continuous-time dynamics
\begin{equation}
	\dot{\mathbf{x}}
	=
	-\mathrm{diag}(\boldsymbol{\mathcal T})\,\mathbf{x}
	+
	\big[\mathbf{W}_S + \mathbf{W}_D(\mathbf{u})\big]\Phi(\mathbf{x})
	+
	\mathbf{I}_S + \mathbf{I}_D(\mathbf{u}),
\end{equation}
where $\boldsymbol{\mathcal T}\in\mathbb{R}^N$ is a learnable neuron-wise leak vector, $\mathbf{W}_S$ and $\mathbf{I}_S$ are static components, and $\mathbf{W}_D(\mathbf{u})$ and $\mathbf{I}_D(\mathbf{u})$ are dynamic components determined by the retrieval cue $\mathbf{u}$ (context).
The static synaptic matrix $\mathbf{W}_S$ is constrained to be symmetric.
After each gradient update, symmetry is enforced by projection $\mathbf{W}_S \leftarrow (\mathbf{W}_S + \mathbf{W}_S^\top)/2$, ensuring that, for any fixed cue, the resulting dynamics remain an attractor system with a nonincreasing energy.

\paragraph*{Cue-dependent synaptic modulation}
In this work, cue-dependent synaptic modulation is parameterized in a simple low-rank form,
\begin{equation}
	\mathbf{W}_D(\mathbf{u})
	=
	(\mathbf{u}\mathbf{W}_{\mathrm{wcue}})^{\top}
	(\mathbf{u}\mathbf{W}_{\mathrm{wcue}}),
\end{equation}
where $\mathbf{W}_{\mathrm{wcue}}$ is a learnable matrix.
This construction ensures that $\mathbf{W}_D(\mathbf{u})$ is symmetric and positive semidefinite for any cue, enabling cue-conditioned reshaping of the energy landscape while preserving pairwise connectivity and avoiding explicit many-body interactions.

The dynamic bias term is defined as
\begin{equation}
	\mathbf{I}_D(\mathbf{u}) = \mathbf{u}\mathbf{W}_{\mathrm{icue}},
\end{equation}
with $\mathbf{W}_{\mathrm{icue}}$ learned from data.
We emphasize that this parameterization is chosen for simplicity and interpretability in the present experiments.
More generally, both $\mathbf{W}_D(\mathbf{u})$ and $\mathbf{I}_D(\mathbf{u})$ can be instantiated as arbitrary functions of the cue, including nonlinear mappings or neural networks, provided that symmetry constraints required for attractor dynamics are respected.

\paragraph*{Energy function}
For a fixed retrieval cue $\mathbf{u}$, DMHN admit an effective energy function
\begin{align}
	E(\mathbf{x})
	=
	&-\frac{1}{2}\Phi(\mathbf{x})^{\top}
	\big(\mathbf{W}_S + \mathbf{W}_D(\mathbf{u})\big)
	\Phi(\mathbf{x})
	\nonumber\\
	&- \Phi(\mathbf{x})^{\top}
	\big(\mathbf{I}_S + \mathbf{I}_D(\mathbf{u})\big)
	+
	\boldsymbol{\mathcal T}^{\top}
	\int_{0}^{\mathbf{x}} \mathbf{v}\,\Phi'(\mathbf{v})\,\mathrm{d}\mathbf{v},
\end{align}
which decreases monotonically along system trajectories for the given cue.
We emphasize that this energy function is defined conditionally on the retrieval cue and characterizes the dynamics within a single retrieval episode.
Different cues thus induce different effective energy landscapes, leading to cue-dependent attractor manifolds rather than a single globally fixed manifold.

\subsection*{Experimental datasets and settings}

We evaluate associative retrieval under a range of synthetic and natural memory settings designed to probe capacity, robustness, and statistical heterogeneity (Fig.~\ref{fig:retrieval}(a)).

\paragraph*{Balanced benchmark patterns}
Balanced benchmark patterns are binary vectors $\mathbf{y}\in\{-1,1\}^N$ with equal activation probability ($p=\frac{\#1}{\#1+\#-1}=0.5$).
Corrupted cues are generated by random bit flips, placing CHN near their theoretical capacity limit ($0.138N$) \cite{HopfieldCapacity}.

\paragraph*{Multi-level imbalanced binary patterns}
To assess robustness to heterogeneous activation statistics, binary patterns are generated with varying activation ratios
$p=\frac{\#1}{\#1+\#-1}\in\{0.1,0.2,0.3,0.4,0.5\}$.
Cues are corrupted using either random masking or additive Gaussian noise, with mask ratios $r$ or noise levels $\sigma$ sampled from a half-normal distribution $\mathcal{N}^+(0,0.3)$.

\paragraph*{Structured binary patterns (MNIST)}
MNIST \cite{MNIST} digits are converted into binary patterns by thresholding grayscale intensities.
These patterns exhibit heterogeneous activation statistics and strong spatial correlations.
Corruption follows the imbalanced binary protocol, using either random masking or additive Gaussian noise.

\paragraph*{Continuous-valued image patterns (CIFAR10)}
For continuous-valued associative memory, CIFAR10 \cite{CIFAR10} images are mapped into a latent space using an encoder network.
Associative retrieval is performed by recurrent Hopfield dynamics in the latent space, and retrieved states are decoded back to image space for evaluation.
Patch-wise masking or additive Gaussian noise is applied to generate corrupted cues.

\subsection*{Training and retrieval protocol}

MHN and DMHN are trained using paired cue-target data $\{(\mathbf{u}, \mathbf{y})\}$ under a unified associative retrieval framework (Fig.~\ref{fig:retrieval}(b)).
For all experiments, retrieval cues are held fixed throughout each retrieval episode, and learning is performed by backpropagation through time.

\paragraph*{Binary pattern retrieval}
For binary pattern datasets, target patterns $\mathbf{y}$ are corrupted to generate cues $\mathbf{u}$.
The network state is initialized as $\mathbf{x}(0)=\mathbf{u}$ and evolved for $T=10$ steps under continuous Hopfield dynamics.
A retrieval loss is computed between the dynamic state and the target pattern at each time step,
\begin{equation}
	\mathcal{L}_{\mathrm{retrieval}}
	=
	\frac{1}{T}
	\sum_{t=1}^{T}
	\left(
	\|\mathbf{x}(t)-\mathbf{y}\|_1
	+
	\|\mathbf{x}(t)-\mathbf{y}\|_2^2
	\right),
\end{equation}
which encourages convergence toward the correct attractor.

Random binary pattern experiments use network dimension $N=100$.
For structured binary patterns derived from MNIST, the network dimension is set to $N=784$, corresponding to the number of pixels.
In both cases, training follows the same binary retrieval protocol, and models are optimized for $10{,}000$ steps.

\paragraph*{Continuous-valued image retrieval}
For continuous-valued memories derived from CIFAR10, an encoder-decoder architecture is introduced to map images to and from a continuous latent attractor space.
The encoder $\mathcal{E}$ and decoder $\mathcal{D}$ are implemented as two-layer perceptrons with GELU nonlinearities and a $\tanh$ output, mapping between image space and an $N$-dimensional latent representation, with $N=1000$.

Both the target image $\mathbf{y}$ and the corrupted cue $\mathbf{u}$ are first encoded into the latent space.
Retrieval dynamics are then performed in the latent space by initializing the network as $\mathbf{x}(0)=\mathcal{E}(\mathbf{u})$ and evolving for $T=10$ steps.
In contrast to the binary setting, the retrieval loss is computed by comparing the network state to the encoded target representation,
\begin{equation}
	\mathcal{L}_{\mathrm{retrieval}}
	=
	\frac{1}{T}
	\sum_{t=1}^{T}
	\left(
	\|\mathbf{x}(t)-\mathcal{E}(\mathbf{y})\|_1
	+
	\|\mathbf{x}(t)-\mathcal{E}(\mathbf{y})\|_2^2
	\right),
\end{equation}
ensuring that associative retrieval is governed by recurrent dynamics in latent space rather than pixel-wise reconstruction.

To support representation learning, a reconstruction loss is applied to both cues and targets,
\begin{align}
	\mathcal{L}_{\mathrm{recon}}
	=
	&\;\|\mathcal{D}(\mathcal{E}(\mathbf{u}))-\mathbf{u}\|_1
	+
	\|\mathcal{D}(\mathcal{E}(\mathbf{u}))-\mathbf{u}\|_2^2
	\nonumber\\
	&+
	\|\mathcal{D}(\mathcal{E}(\mathbf{y}))-\mathbf{y}\|_1
	+
	\|\mathcal{D}(\mathcal{E}(\mathbf{y}))-\mathbf{y}\|_2^2 .
\end{align}
The total training objective is given by
\begin{equation}
	\mathcal{L}
	=
	\mathcal{L}_{\mathrm{retrieval}}
	+
	\lambda\,\mathcal{L}_{\mathrm{recon}},
\end{equation}
with $\lambda=5$.
Models trained on CIFAR10 are optimized for $20{,}000$ steps.

\paragraph*{Optimization and parameter initialization}
All models are optimized using AdamW with learning rate $5\times10^{-4}$, $\beta_1=0.9$, $\beta_2=0.95$, and weight decay $10^{-2}$.
An exponential learning rate scheduler with decay factor $\gamma=0.9$ is applied every 500 optimization steps.

All synaptic weights are initialized from a uniform distribution $\mathcal{U}(-1/\sqrt{N},\,1/\sqrt{N})$.
Leak and bias parameters are initialized to zero.
After each optimization step, symmetry of the static Hopfield weight matrix is enforced by projection,
and weights are rescaled to maintain a bounded spectral norm.
Leak parameters are constrained to remain nonnegative throughout training.

\subsection*{Metrics}

During evaluation, all parameters are held fixed.
Given a cue, the network is evolved under continuous dynamics until convergence, yielding a cue-conditioned attractor state that is used to assess retrieval accuracy.

\paragraph*{Binary pattern retrieval}

For binary pattern datasets, retrieval performance is evaluated by comparing the thresholded final network state $\mathrm{sgn}(\mathbf{x}(T)) \in \{-1,1\}^N$ with the target pattern $\mathrm{sgn}(\mathbf{y}) \in \{-1,1\}^N$.
The bit-wise error rate is defined as
\begin{equation}
	\mathrm{err}
	=
	\frac{1}{N}
	\sum_{i=1}^{N}
	\mathbb{I}\!\left[
	\mathrm{sgn}\!\left(x_i(T)\right)
	\neq
	\mathrm{sgn}\!\left(y_i\right)
	\right],
\end{equation}
where $\mathbb{I}[\cdot]$ denotes the indicator function.
A retrieval is considered correct if $\mathrm{err}$ falls below a predefined threshold.

Retrieval quality is additionally characterized using the overlap between the retrieved state and the target pattern, defined as
\begin{equation}
	m
	=
	\frac{1}{N}
	\sum_{i=1}^{N}
	\mathrm{sgn}\!\left(x_i(T)\right)\,\mathrm{sgn}\!\left(y_i\right).
\end{equation}
For binary patterns, the overlap $m$ is linearly related to the bit-wise error rate via $m = 1 - 2\,\mathrm{err}$ and provides an equivalent but geometrically interpretable measure of alignment between the retrieved state and the stored pattern.
Overlap distributions are used to analyze the stability and separability of attractor states across models.

Given differences in task difficulty across settings, retrieval thresholds are specified accordingly and held fixed across models.
For balanced benchmark datasets with flip noise ($p=0.5$), a retrieval is deemed correct when $\mathrm{err} < 5\%$.
For all other binary pattern settings, including imbalanced activation statistics and alternative corruption models, a stricter criterion of $\mathrm{err} < 1.5\%$ is used.
These thresholds are held fixed across models and reflect differences in baseline difficulty across settings.

\paragraph*{Continuous-valued memory retrieval}
For continuous-valued memories derived from CIFAR10, retrieval performance is quantified using the mean squared error (MSE) between the decoded network state $\mathcal{D}(\mathbf{x}(T))$ and the target image $\mathbf{y}$:
\begin{equation}
	\mathrm{MSE}
	=
	\frac{1}{HWC}
	\left\|
	\mathcal{D}(\mathbf{x}(T)) - \mathbf{y}
	\right\|_2^2,
\end{equation}
where $H$, $W$, and $C$ denote the image height, width, and number of channels, respectively.
A retrieval is considered correct if $\mathrm{MSE} < \frac{25}{HWC}$, corresponding to an average per-pixel error of $\sim0.8\%$ for $32\times32\times3$ CIFAR10 images.

All reported metrics are averaged over 4 random initializations, 4 noise realizations, and independent trials.
These evaluation criteria directly correspond to the performance measures reported in Fig.~\ref{fig:retrieval}, Fig.~\ref{fig:pattern-results}, and Tab.~\ref{tab:retrieval-results}.

\section*{Data, Materials, and Software Availability} The data and code supporting the findings of this study will be made publicly available upon publication.

\section*{ACKNOWLEDGMENTS} This work was supported in part by the 2025 YangtzeRiver Delta Science and Technology Innovation Community Joint Research (Basic Research) Project (Grant No. 2025CSJZN00100), the LingangLaboratory (Grant No. LGL-1987-10), and Shanghai Neuhelium Neuromorphic Intelligence Technology Co., Ltd.

\bibliographystyle{unsrt}  
\bibliography{references}  

\end{document}